# Highlights

**Robust Ensemble Person Re-Identification via Orthogonal Fusion with Occlusion Handling**

Syeda Nyma Ferdous, Xin Li

- Propose a novel ensemble learning-based approach for occluded reID problem.
- Using the orthogonality principle, our developed deep CNN model uses masked autoencoder and global-local feature fusion for occlusion robust person re identification.
- Design a part occluded token-based Transformer classifier to generate occlusion robust embedding.
- Our proposed model leverages the benefits of both Transformer and CNN using ensemble learning.

# Robust Ensemble Person Re-Identification via Orthogonal Fusion with Occlusion Handling

Syeda Nyma Ferdous[a], Xin Li[b]

[a]*Lane Department of Computer Science and Electrical Engineering, West Virginia University, Morgantown, 26506, West Virginia, United States*
[b]*Department of Computer Science, University at Albany, State University of New York, 12222, New York, United States*

**Abstract**

Occlusion remains one of the major challenges in person reidentification (ReID) as a result of the diversity of poses and the variation of appearances. Developing novel architectures to improve the robustness of occlusion-aware person Re-ID requires new insights, especially on low-resolution edge cameras. We propose a deep ensemble model that harnesses both CNN and Transformer architectures to generate robust feature representations. To achieve robust Re-ID without the need to manually label occluded regions, we propose to take an ensemble learning-based approach derived from the analogy between arbitrarily shaped occluded regions and robust feature representation. Using the orthogonality principle, our developed deep CNN model makes use of masked autoencoder (MAE) and global-local feature fusion for robust person identification. Furthermore, we present a part occlusion-aware transformer capable of learning feature space that is robust to occluded regions. Experimental results are reported on several Re-ID datasets to show the effectiveness of our developed ensemble model named orthogonal fusion with occlusion handling (OFOH). Compared to competing methods, the proposed OFOH approach has achieved competent rank-1 and mAP performance.

## 1. Introduction

Research on person re-identification (Re-ID), a problem of matching person images from different cameras, has been active for the past decade (1). Various applications of this problem are gaining interest in academia and industry, including surveillance, security, autonomous driving, and activity analysis (2). There are various obstacles that can obstruct an image in a real-world setting, such as trees, cars, and even other people. ReID poses a critical problem since we need to match images from different angles and perspectives. It is even more difficult by occlusion, leading to the inclusion of less discriminatory information in the image for image matching (3). In addition to person Re-ID, occlusion handling has remained a challenge in human detection (4), scene segmentation (5), object tracking (6) and stereo matching (7).

The existing occlusion handling strategies in the literature on person ReID can be classified into the following categories. First, it is natural to tackle the occlusion problem using the attention mechanism (8) - i.e., to pay more attention to unoccluded regions than to occluded ones. Several works (e.g., (3), (9), (10)) have shown improved robustness to occlusion by designing attention-guided (or pose-guided) loss functions. Second, some work, such as the semantic-aware occlusion-robust network (SORN) (11) and high-order Re-ID (12) addresses the issue of occlusion at the higher semantic level, i.e., both local and global information contained in the occluded images will be extracted and combined adaptively to improve the robustness of the overall system. For the occluded REID problem, the data distribution of occluded and non-occluded samples is significantly different (13). This phenomenon demands a generalizable solution that can articulate the domain shift between source and target data samples. Along this line of reasoning, we propose to adopt an ensemble learning approach using both CNN and Transformer architecture, as shown in Fig. 1. Such a model can improve the ability of feature representation for occlusion handling.

The key challenge in the occluded ReID problem is how to learn discriminative information from occluded data. Also, occluded images lack identity information, which is crucial for designing robust re-identification.

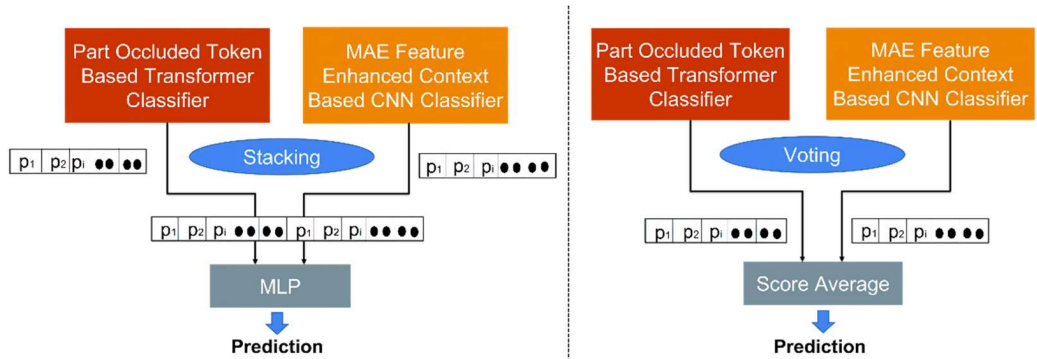

Figure 1: Our proposed ensemble model consists of two complementary components for orthogonal fusion with occlusion handling (OFOH): MAE feature enhanced *context*-based CNN model and partly occluded *token*-based transformer model. The final prediction can be performed by stacked generalization (left) or a voting-based approach (right).

system. A deep model that captures context-based information can be a potential solution to this problem. To address these challenges, we design a pixel context-based CNN model that utilizes both local and global features for occluded RE-ID tasks. To increase identity-related information, we augment the feature space with features extracted from reconstructed images generated by the self-supervised learner, Masked Autoencoder (MAE) (14). To generate a discriminative feature space, we propose a solution leveraging the idea of deep orthogonal fusion of local and global features (DOLG) model (15). The proposed model can combat occlusion interference by adaptively fusing the features of the global and local body parts. Orthogonal fusion preserves critical local information, eliminating redundant information for the fusion of local and global features. To our knowledge, the DOLG model has not been considered for handling occlusion in open literature. The new insight brought by this work is that occlusions must be localized events due to the constraints of physical laws. Recognizing the analogy between image retrieval (the target application of DOLG) and person Re-ID, an ID-preserving person ReID model can be developed utilizing effective local-global feature fusion for occlusion handling. Our second model focuses on achieving invariance to occlusion using body part-occluded tokens. To design a ReID model invariant to occlusion, we propose an orthogonal fusion with occlusion handling (OFOH) model based on Transformer architecture that generates an occlusion-invariant feature space. CNN-based verifier generates a supervisory signal generating part-occluded tokens.

The key technical contributions of this paper are summarized below.

- We propose an approach that uses an ensemble learning paradigm for the occluded person Re-ID problem. For the first time, we advocate the ensemble of CNN and transformer models with complementary properties (i.e., context-based vs. token-based) for occluded ReID.

- We apply orthogonal fusion to our CNN model that helps learn discriminative global and body-part local features that avoid interference with occlusion. Unlike DOLG (15), our orthogonal fusion module combines the local part features extracted from human body parts and ID information enhanced global features.

- Our transformer-based model learns part-occluded tokens for occluded ReID with the supervision of a CNN verifier.

- The experimental results on our OFOH achieve impressive results on both occluded and holistic Re-ID datasets, including Occluded ReID, Occluded DUKE, Market-1501, and DukeMTMC-reID dataset. The proposed OFOH approach has achieved competent and sometimes better rank-1 and mAP performance.

## 2. Related Work

**Person Re-ID**. Person Re-ID has been extensively studied in the literature (1; 16). A recent survey on deep learning for Person Re-ID is referred to (17). Among the most influential works on person Re-ID, the harmonious attention network (18) has shown the advantages of jointly learning attention selection and feature representation in a Convolutional Neural Network (CNN) by maximizing complementary information at different levels of visual attention subject to discriminative learning constraints of Re-ID. More recently, such a joint learning framework was extended by coupling Re-ID learning and data generation end-to-end (i.e., joint discriminative and generative learning (19)). Group-aware label transfer (GLT) (20) was proposed for unsupervised domain adaptive person Re-ID; a similar unsupervised person Re-ID method was presented in (21) that focuses on considering the distribution discrepancy between cameras. A meta-distribution alignment strategy has been proposed for domain generalizable person ReID (22). Inspired by the latest advances in transformers, a new part-aware transformer (PAT) was developed for the occluded person Re-ID in (23) through a transformer encoder-decoder architecture. Most recently, camera-aware representation learning (CARL) has been proposed in (24) for person reid to handle camera biased problems. And cross-modality person Re-ID explores modality-shared appearance features and modality-invariant relation features in (25).

**Occlusion Handling**. In the early work of the HOG-LBP-based human detector, occlusion was handled by a dedicated likelihood map with mean shift segmentation in (4). This part-based tracking with partial occlusion handling was extended to multiple-person tracking in (26) under a tracking by-detection framework. Despite the effectiveness of part-based models, a key issue is how to integrate inaccurate scores from part detectors when there are occlusions or large deformations (27). In (27) a deformable part-based model was developed to obtain the score of the part detectors (the visibility scores of the parts are modeled as hidden variables). A bidirectional aggregation network (BANet) with occlusion handling was proposed for panoptic segmentation (simultaneous instance and semantic segmentation) in (28). a novel occlusion handling strategy was proposed in (29) to explicitly model the relationship between occluding and occluded tracks, outperforming previous part-based approaches. A novel feature erasing and diffusion network has been developed for occluded ReID in (30). Most recently, a body part based ReID model has been proposed in (31) using GiLT, a novel training strategy.

**Ensemble Learning**. Ensemble learning (32) is a classical machine learning approach that tends to combine the benefits of multiple models (33). In (34), an ensemble model was developed for person re-identification (EnsembleNet), which effectively learns feature representation of individual persons for saving computation and memory cost. EnsembleNet constructed a feature space by dividing a single network (ResNet-50) into multiple branches and concatenating each of the branch features into a single vector to represent a person. For person Re-ID, an end-to-end ensemble learning method was proposed for discriminative feature learning in (35) to address the overfitting problem. A self-ensemble learning has been proposed for the unsupervised domain-adaptive person ReID problem (36), which is based on self-ensemble prediction to increase the quality of mean teacher-based soft labels. More recently, a local correlation ensemble model has been applied to cross-domain Re-ID in (37).

**3. Methodology**

*3.1. System Overview*

In this paper, we propose a deep ensemble learning framework for occlusion handling for person re-identification. Our approach learns a robust latent representation for the occluded person Re-ID problem using two deep models. Our first model utilizes the benefits of self-supervised models, Masked Autoencoder (MAE) (14) by improving the features of its backbone with features of reconstructed masked images to handle occlusion.

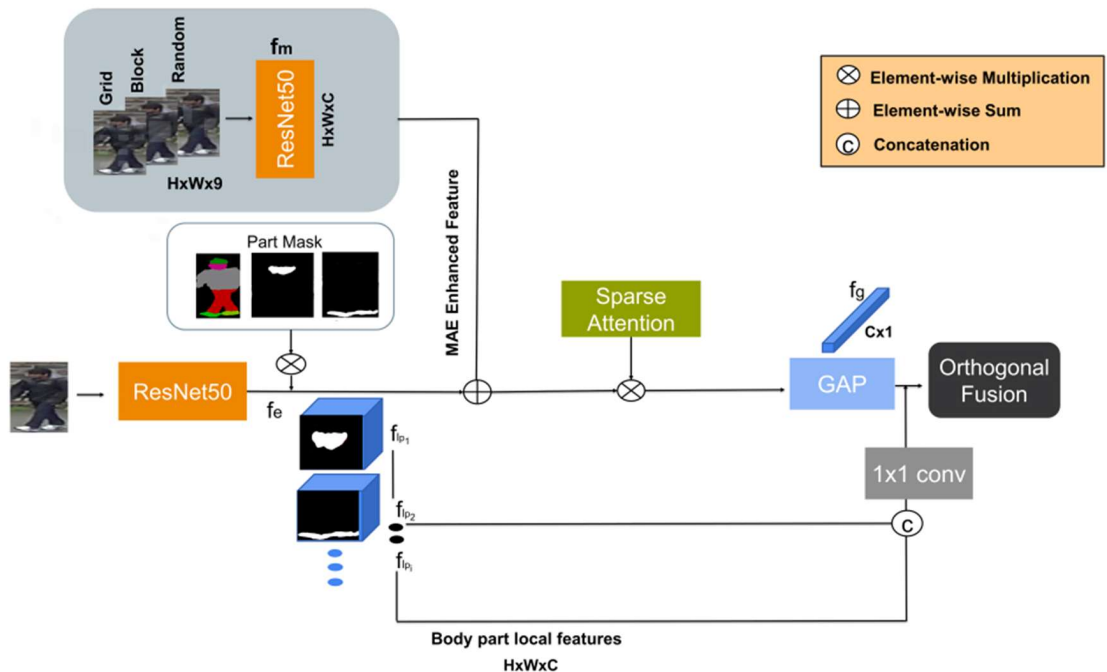

Figure 2: Architecture of our proposed Context Based CNN Classifier. First, we extract local body features using a body part mask. To generate an occlusion-robust global representation, features from Masked Autoencoder (MAE) reconstructed images are utilized. Induced noise in MAE enhanced feature is suppressed using sparse attention. For training, part mask guided local features along with occlusion-robust MAE enhanced global features are concatenated using Orthogonal fusion.

Our second model leverages the benefits of transformer architecture with a carefully designed occlusion handling pipeline. Figure 1 illustrates an overview of our proposed approach. Our system design is based on the following motivating observations.

1. Occlusion vs. Masking. The analogy between the phenomenon of occlusion in the physical world and the masking strategy in the virtual world is apparent. Such an analogy has inspired the study of robust face recognition with occlusion in the literature (38; 39). In contrast, their subtle difference has been under-researched. From the perspective of the simulation-to-real gap (40), we observe that occlusions in the physical world (e.g., a wearing hat and a blocking bush) are characterized by the irregular geometry of the object in the foreground. Such a lack of geometric regularity in occluded objects in the real world is strikingly different from the use of regular masks (e.g., rectangular-shaped) in data augmentation studies (e.g., CutMix (41)).
2. Identification vs. Reconstruction. Reconstructed images as data augmentation have already been explored and found to be effective for the ReID problem. For example, CamStyle (42), PTGAN (43) GAN-based models to augment training dataset for person ReID. The success of applying masked image modeling (e.g., MAE (44)) as a robust vision learner, has inspired us to utilize the features from MAE reconstructed images as a strategy to generate a robust feature space.

3. Local vs. Global Features. Recognizing that occlusion is a localized phenomenon, an effective occlusion handling strategy can be developed that combines global and local characteristics (e.g., the celebrated HOG-LBP human detector (4)). Recent advances such as Deep Orthogonal Local and Global (DOLG) (15) have also shown the benefit of combining global and local features by the orthogonality principle for image retrieval applications. However, it is unclear how the desirable orthogonality constraint can be satisfied in the presence of occlusion.

*3.2. Masked Autoencoder (MAE) Feature Enhanced Context based CNN Classifier*

The architecture of the proposed pixel context-based CNN classifier is depicted in Fig. 2.

3.2.1. *Masked Autoencoder (MAE) Enhanced Feature Generation*

To model the image context information, we adopt a pixel-context-based approach for occluded person reidentification. The proposed model uses ResNet-50 (45) without global average pooling and a fully connected layer as a backbone for feature extraction. We combine the benefits of masked autoencoder (MAE) (14) to learn robust visual representation. MAE is a very recent self-supervised pre-training paradigm that follows a simple but effective way to solve computer vision problems such as image reconstruction, image classification, object detection, and others. MAE only needs partial information to reconstruct the original signal. The reconstructed images are generated from different masked versions of the original image which can aid in learning an occlusion-invariant feature representation. We generate MAE reconstructed images applying different mask sampling strategies (e.g., block, grid, random) as shown in Fig. 3 and concatenate these reconstructed images in a channel-wise fashion.

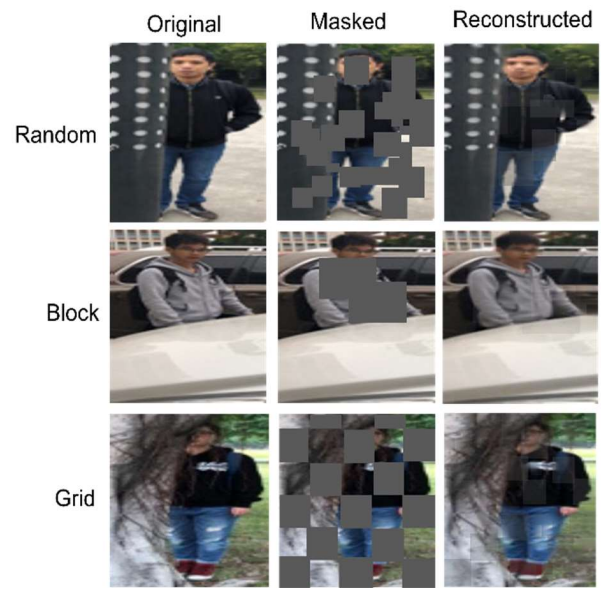

Figure 3: MAE reconstruction results for different mask sampling strategies (Random, Block and Grid)

Then, we extract features from the concatenated image of size $H \times W \times 9$ using a CNN backbone, in our case, ResNet-50 (45). We augmented the feature space of our developed model using the features extracted from MAE reconstructed images to form an id-preserving occlusion noise-robust feature map representation $H \times W \times C$.

*3.2.2. Sparse Attention (SA)*

Attention (46) is used to model the importance of pixels in a feature map. For an input feature map $x \in H \times W \times C$, attention is calculated using the query $q$, the key $k$, and the value $v$ feature embedding as in Eq. (1).

$$s_{ij} = (\frac{qk^T}{\sqrt{d_k}})v, \tag{1}$$

i and j denote the index of the query and the key positions in an embedding space, and $d_k$ is the dimension of the query and the key. As mentioned above, we have augmented the feature space of the original image with a concatenated MAE reconstructed image. Although MAE-reconstructed images help generate a discriminative feature space that is aware of occlusion, this might induce noise in the system as the distribution of MAE-reconstructed images is different from the original ones. To suppress noise, we use sparsemax (47) instead of the softmax function while computing attention in the feature space.

$$softmax_{(s_{ij})} = \frac{exp(s_{ij})}{\sum_{j=1}^{hw} exp(s_{ij})}, \tag{2}$$

$$sparsemax_{(s_{ij})} := argmin_{p \in \Delta^{K-1}} ||p - s_{ij}||^2, \tag{3}$$

where $\Delta^{K-1} := \{p \in \mathsf{R}^K | 1^T p = 1, p \geq 0\}$ be the ($K-1$)-dimensional simplex.

*3.2.3. Global and Local Feature Concatenation using Orthogonal Fusion*

For occluded person Re-ID, we need discriminative features from body parts along with global features. Despite the popularity of using masks in person re-identification, how to generate context-dependent masks has remained under-researched. In existing mask generation methods (e.g. CutMix (41)), the mask shape is constrained to be rectangular. By contrast, occlusion in the physical world is often characterized by irregular shapes of moving objects (e.g., a wearing hat or a blocking tree). In the context of person Re-ID, we argue that generating human-oriented masks from real-life aspects is desirable because it will guide model training to consider only important regions eliminating the background. To reduce the effect of background clutters and help the model learn body part features, we utilize masks generated by the Self-Correction for Human Parsing (SCHP) (48) model. For each part identified in the generated segmentation mask, we generate a binarized map and multiply the map by the extracted feature map to generate local features of the body part $f_{lp_i}$. This helps the model to focus on visible body parts. Finally, all local features of body parts are concatenated and processed by a 1×1 convolution layer.

To obtain a global feature representation, we concatenate the feature map extracted from the original image and the MAE-reconstructed image and perform sparse attention followed by Global Average Pooling (GAP) and get $f_g \in R^{1 \times C}$. It is natural to think of a solution that considers both local and global features jointly. Thanks to orthogonal fusion, deep features are efficiently fused using orthogonal projection (15; 49). In this paper, we perform an orthogonal fusion of the features of global and local parts following the DOLG model (15). Orthogonal fusion concatenates the global and local features of an instance in orthogonal space so that the most critical local information is preserved while the redundant global information is eliminated. To perform orthogonal fusion of global and local features, local feature projection $f_{lproj}$ is computed.

$$f_{lproj} = \frac{f_l \cdot f_g}{|f_g|^2} f_g \qquad (4)$$

Then, the orthogonal component of each feature point is computed by calculating the subtraction of local feature and its projection vector. Finally, we concatenate the projected local orthogonal features with the global features. A fully connected layer is used to compute the final C×1 descriptor.

### 3.3. Part Occluded Token-based Transformer Classifier

Our second model is based on the Transformer architecture, ViT (50) as shown in Fig. 4. ViT divides an input image of size $\forall x \in R^{H \times W \times C}$ into non-overlapping patches of equal size to get a flattened 1D vector representation. To be more specific, ViT divides a 2D image into *PxP* flattened patches where $\forall x \in R^{N \times P^2 \times c}$ where N stand for the number of total patches and C stands for the number of channels.

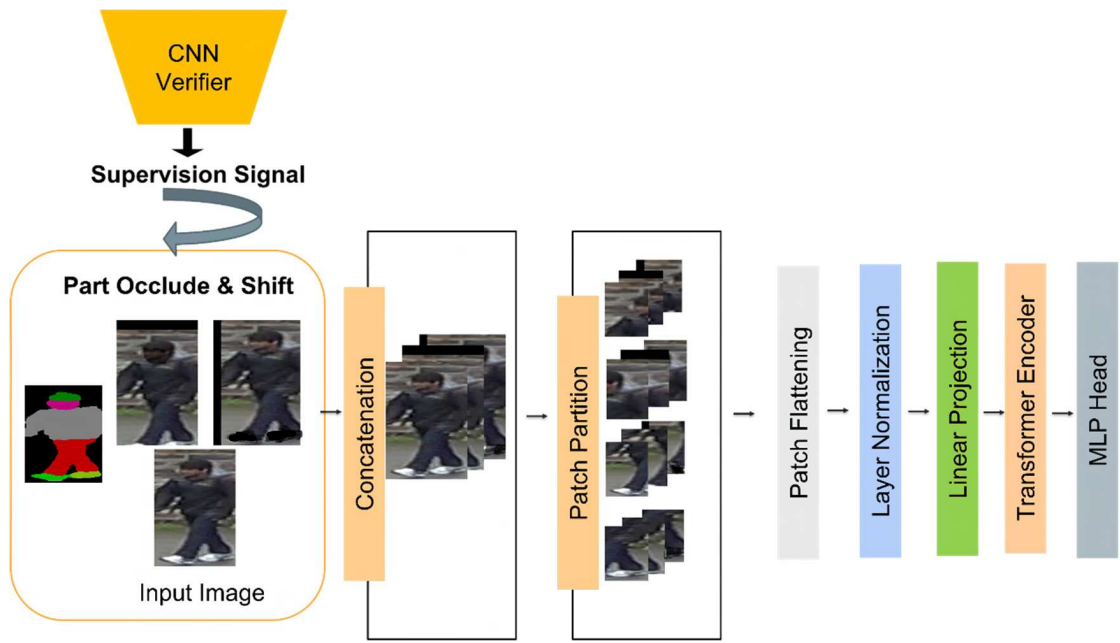

Figure 4: Architecture of our proposed occlusion robust Transformer classifier. First, part mask is used to generate part occluded samples. Then, the verifier selects discriminative part occluded samples, and the selected samples are concatenated and flattened after patch partitioning. Each concatenated patch is linearly projected and fed to the Transformer encoder. The classification is performed with MLP head.

To make our model robust to occlusion, we modified the input processing of the original ViT. The proposed modification is derived from the design of the Shifted Patch Tokenization (SPT) module (51). SPT effectively increases the locality inductive bias of the original ViT model. Inspired by this, we propose an occlusion robust version of ViT that considers part-occluded tokens together with the original tokens. First, we mask out parts from images to generate part-occluded images under the guidance of masks generated from the Self-Correction Human Parsing (SCHP) model (48). To select part-occluded images close to the original ones, we take the help of a Siamese CNN verifier. The backbone of the verifier is MAE Feature Enhanced Context-based model. We also perform random shift operations on the part-occluded images to add inductive bias to the model. Next, we concatenate the selected part-occluded images with the original input image. The concatenated image is divided into non-overlapping patches and then the patches are flattened. Layer Normalization (LN) and Linear Projection are applied on the flattened patches to obtain a sequence of visual tokens. Each patch embedding corresponds to a visual token which is the input to the Transformer encoder. This process is called Tokenization which converts the 3D feature tensor, $\forall x \in R^{H \times W \times C}$ into 2D matrix feature, $\forall x \in R^{N \times d}$ *where N* represent the number of patches and d represent hidden dimension of transformer encoder. An extra classification token, $x_{cls}$ containing the representation information for the entire image, is attached to the visual tokens. Then, a learnable positional embedding is added to provide positional information to the original embedding. The input sequences fed to the transformer can be expressed using Eq. (4).

$$I_e = [x_{cls};\ LN\ (P([x\ o^1 o^2 ... o^N])\ E] + E_{pos}, \tag{4}$$

Here, $o_i \in R^{H \times W \times C}$ represents the $i^{th}$ part-occluded images, $E \in R^{(P^2 \cdot C \cdot (N+1) \times D)}$ is a learnable linear projection of tokens, $E_{pos} \in R^{(N+1) \times D}$ denotes a learnable positional embedding, $N$ is the number of tokens embedded and $D$ represents the hidden dimension of the transformer encoder.

The transformer encoder consists of two main components: multi-head self-attention (MHSA) and multi-layer perceptron (MLP). Finally, the classification is performed by the MLP head. The MLP contains two layers with a GELU non-linearity. Multi-head self-attention (MHSA) expands upon single-head self-attention by incorporating complex connections between different image patches. The self-attention mechanism serves as a foundational component of Transformers, systematically modeling the interactions among all pixels in an image by assigning attention scores to pairs of patch tokens based on global contextual information. The classification is performed by MLP head.

*3.4. Ensemble Model of Orthogonal Fusion with Occlusion Handling (OFOH)*

To construct a deep ensemble model, we incorporate two techniques: voting and stacking. In the voting approach, we average the prediction scores of independent models. To implement a stacked generalization model, we train an MLP as a Meta-Learner where we first concatenate the prediction of individual models and then train an output fully connected, fc layer for predicting person ID.

*3.5. Overall Training Objective*

To design a Re-ID classifier that preserves discriminative information, we train our model with cross-entropy loss, such as ID loss and triplet loss (52). ID loss is computed using the following formula:

$$L_{id(f^*)} = -\lambda_{id} \log(p) \qquad (5)$$

Here, $f^*$ is the global feature representation, $p$ is the probability predicted by the identity classifier. The triplet loss is calculated as follows.

$$L_{tri(f^*)} = \alpha \, [d(f^*, f_p^*) - d(f^*, f_n^*)] \tag{6}$$

$\alpha$ is the margin for triplet loss, $d(f^*, f_p^*), d(f^*, f_p^*)$ are the distances calculated for anchors from positive and negative samples.

The overall classification loss is computed using Eq. (7).

$$L_{dis(f^*)} = L_{id(f^*)} + L_{tri(f^*)} \tag{7}$$

To train our stacked generalization model, we try to reduce redundancy among multiple model predictions by applying diversity regularization (53) on the fully connected layer of MLP. This regularization ensures diverse weight vectors by maximizing the minimal pairwise angles between vectors. Then, the classification loss can be formulated as follows.

$$L_{Total} = L_{dis(f^*)} + \lambda L_{div}(W_{classifier}) \tag{8}$$

where $\lambda$ is the hyperparameter associated with diversity loss.

$$L_{div} = -\frac{1}{n}\sum_{i=1}^{n} \min_{j, j \neq i} \theta_{ij}, \ \theta = arccos(\tilde{W}\tilde{W}^T), \ \tilde{W}_i = \frac{W_i^T}{||W_i||} \tag{9}$$

**4. Experimental Results**

*4.1. Dataset and Implementation Details*

To understand the performance of our proposed model, we performed our experiments on occluded and partial reid datasets, namely Occluded-

REID (10) and Partial REID (54). We also carried out our experiment on two holistic Re-ID datasets, Market-1501 (16) and DukeMTMC-reID (55). To further verify the robustness of our model in aerial surveillance, we performed our experiment on the PRAI-1581 dataset (56). The details are given below.

**Occluded-REID**: Occluded-REID (3) consists of 2,000 images of 200 occluded persons. Each identity has five full-body person images and five occluded person images with different types of severe occlusions.

**Occluded-Duke**: Occluded-Duke (10) is derived from DukeMTMC-reID dataset. The training set contains 15,618 images of 720 people and the testing set contains 17,661 gallery images and 2,210 query images of 1,100 people.

**Market-1501**: Market-1501 dataset (16) consists of 1,501 different identities captured by 6 cameras. The training, query, and gallery set consist of 12,936, 3,368 and 19,732 images, respectively.

**DukeMTMC-reID**: DukeMTMC-reID (57) contains 36,411 images of 1,404 identities captured by 8 cameras. The training, gallery, and query set consist of 16522, 17661, and 2228 images, respectively.

**PRAI-1581**: PRAI-1581 (56) is a fairly recent publicly released dataset for research purposes in aerial surveillance. This dataset contains 39,461 person images of 1581 classes captured by two UAV drones with a flight altitude ranging from 20 to 60 meters above the ground. Due to the wide range of flight altitudes, the pose variations and occlusion of this data set are noticeably more challenging than those used in normal surveillance Re-ID research.

DEM1 denotes proposed Context Based CNN Classifier while DEM2 represents part occluded token-based Transformer classifier. DEMV and DEMS denote Deep ensemble-based voting and stacking model for occluded person reidentification. To train the context-based CNN classifier (DEM1), we resize all input images to 256×128. The batch size is set to 64. The model is trained for 150 epochs. The learning rate is initialized to $2.5 \times 10^{-4}$ and decayed to its 0.1 and 0.01 at 30 and 90 epochs. The training images are augmented with random cropping and random erasing. For training occlusion robust Transformer classifier (DEM2), the depth was set to 9, hidden dimension was set to 199. To generate MAE reconstructed images, we fine-tune the pre-trained MAE model (ViT-L). Figure 3 visualizes the different masks and corresponding images generated from MAE. The batch size is set to 64. The initial learning rate is set to 0.008 with the cosine learning rate decay. The model is trained with AdamW optimizer (58). Note that we start with the initial pre-trained weights of ImageNet1k (IN1K) and later fine-tune them for person Re-ID datasets. The model is trained for 350 epochs.

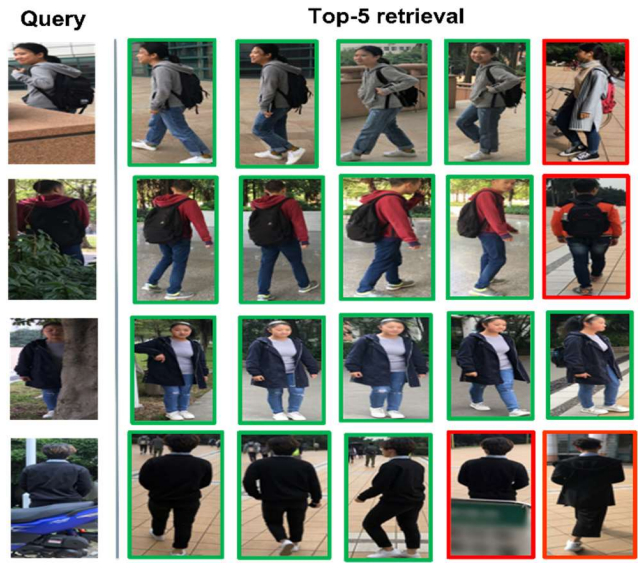

Figure 5: Top-5 ranking results for Occluded-REID dataset using our method. Red rectangular boxes indicate false image retrieval, while green boxes indicate right ones.

Top-5 retrieval results using our best-performing deep ensemble method are shown in Fig. 5. Grad-CAM visualization (59; 60) of ensemble models is shown in Fig. 6. Both models focus on distinctive body part features while suppressing occluded regions in the image. The visualization results indicate that our proposed approach can successfully handle person re-identification in the presence of occlusion.

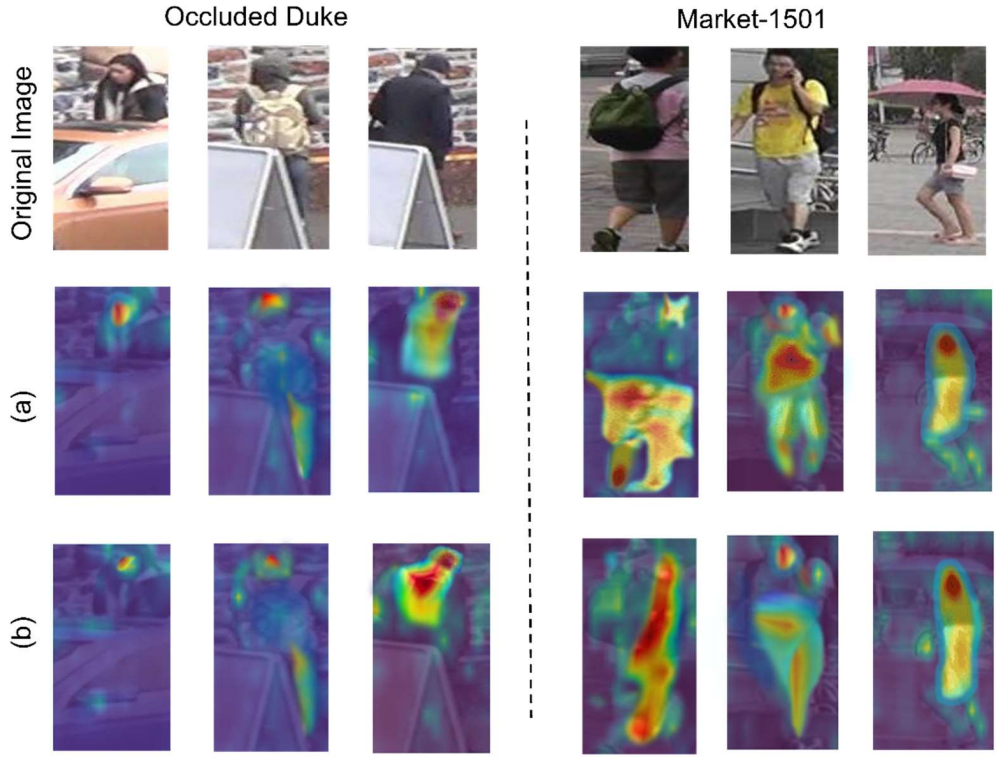

Figure 6: Grad-cam visualization for Occluded Duke and Market-1501 dataset. (a) DEM1 (b) DEM2.

## 4.2. Comparison with State-of-the-Art Methods

We compare our method with previous person Re-ID approaches. To obtain a fair comparison, all methods are compared in terms of the rank-1 accuracy (R1) and mean average precision (mAP) metric. In Table 1, our proposed method is compared for the occluded REID and Occluded-DukeMTMC datasets. Our best-performing model outperforms previous state-of-the-art methods for the occluded ReID and Occluded-DukeMTMC datasets demonstrating the effectiveness of our method for handling occlusion.

Table 1: Performance comparison for Occluded REID and Occluded Duke.

| Method | Occluded REID | | Occluded Duke | |
|---|---|---|---|---|
| | R1 | mAP | R1 | mAP |
| PCB (61) | 41.3 | 38.9 | 42.6 | 33.7 |
| FPR (62) | 78.3 | 68.0 | - | - |
| PVPM (63) | 70.4 | 61.2 | 47.0 | 37.7 |
| HOReID (12) | 80.3 | 70.2 | 55.1 | 43.8 |
| PAT (23) | 81.6 | 72.1 | 88.0 | - |
| TransReID (64) | - | - | 66.4 | 59.2 |
| PFD (65) | 81.5 | 83.0 | 69.5 | 61.8 |
| QPM *(66) | - | - | 66.7 | 53.3 |
| **OFOH(Ours)** | **81.9** | **83.1** | **71.7** | **62.6** |

Table 2: Performance comparison for Market-1501 and DukeMTMC-REID.

| Method | Market-1501 | | DukeMTMC-reID | |
|---|---|---|---|---|
| | R1 | mAP | R1 | mAP |
| PCB (61) | 92.3 | 77.4 | 81.8 | 66.1 |
| MGN (67) | 95.7 | 86.9 | 88.7 | 78.4 |
| HOReID (12) | 94.2 | 84.9 | 86.9 | 75.6 |
| IANet (68) | 94.4 | 83.1 | 87.1 | 73.4 |
| PGFA (10) | 91.2 | 76.8 | 82.6 | 65.5 |
| PAT (23) | 95.4 | 88.0 | 88.8 | 78.2 |
| TransReID (64) | 95.2 | 88.9 | 90.7 | 82.0 |
| EnsembleNet (34) | 95.6 | 93.0 | 90.1 | 88 |
| PFD (65) | 95.5 | 89.7 | 91.2 | 83.2 |
| **OFOH(Ours)** | **95.6** | **89.6** | **91.4** | **83.3** |

As shown in Table 1, the rank-1 results on occluded REID and Occluded-DukeMTMC datasets are 81.9% and 88.7% which outperform the rank-1 accuracy of previously reported person Re-ID methods. The reason may be efficient part learning in the presence of occlusion using our proposed model. The results in Table 1 ensure that our model is robust to occlusion. We also apply our method to holistic Re-ID datasets. In Table 2, we report the results for the Market-1501 and DukeMTMC-reID datasets. The proposed method achieves competitive performance compared to state-ofthe-art methods. Specifically, the rank-1/mAP for Market-1501 is 95.6%/ 89.6 and for DukeMTMC-reID 91.4%/ 83.3%.

Table 3: Performance comparison for PRAI-1581.

| Method | PRAI-1581 | |
| --- | --- | --- |
|  | R1 | mAP |
| PCB (61) | 47.47 | 37.15 |
| MGN (67) | 49.64 | 40.86 |
| OSNet (69) | **54.40** | **42.10** |
| **OFOH(Ours)** | 53.91 | 41.22 |

To further assess the performance of our model, we experiment with our method for aerial surveillance. We performed the experiment on the PRAI dataset. From the results in Table 3, we can see that the performance of our proposed method is highly comparable to the state-of-the-art in aerial surveillance. CMC curves for the studied datasets are shown in Fig. 7. The results show that deep ensemble models perform better compared to individual models for the studied ReID datasets.

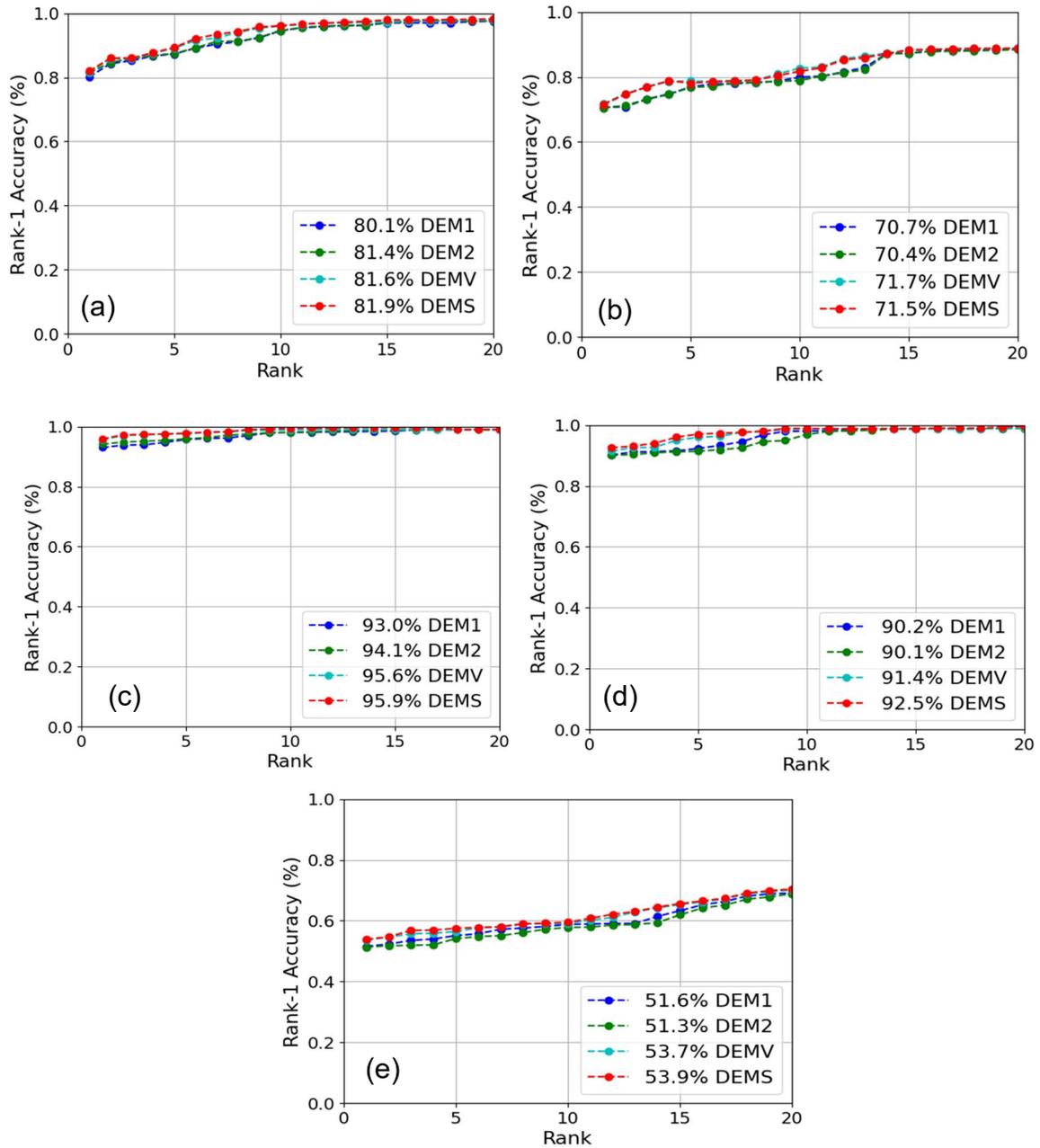

Figure 7: CMC Curves for both independent (DEM1- Context Based CNN Classifier, DEM2- part occluded token-based Transformer classifier) and deep ensemble models (DEMV- Deep ensemble-based voting, DEMS- Deep ensemble based stacking) for different datasets (a) Occluded REID (b) Occluded Duke (c) Market-1501 (d) DukeMTMC-reID and (e) PRAI-1581

*4.3. Ablation Studies*

In this section, we report several ablation studies on Occluded-REID dataset using our proposed approach.

**Effectiveness of MAE Reconstructed Feature Fusion:** The effectiveness of MAE-reconstructed feature fusion is shown in Table 4. The fusion of MAE-reconstructed features attains a performance increase, improving rank1 accuracy by +1.9% and mAP by +1.4%.

**Table 4:** Effect of MAE reconstructed feature fusion while training DEM1. The backbone of the ensemble model is ResNet-50.

| Module | Rank-1 | mAP |
| --- | --- | --- |
| DEM1 w/o MAE enhanced feature | 78.2 | 81.4 |
| DEM1 w/ MAE enhanced feature | 80.1 | 82.8 |

**Impact of hyper-parameter, $\lambda$:** To optimize the training of the deep ensemble model, we set the hyper-parameter, $\lambda$ to tune the impact of weight diversity in both models. The result of varying $\lambda$ is shown in Fig. 8. For the best-performing deep ensemble model, in this case, stacking is attained by setting the $\lambda$ to 0.01.

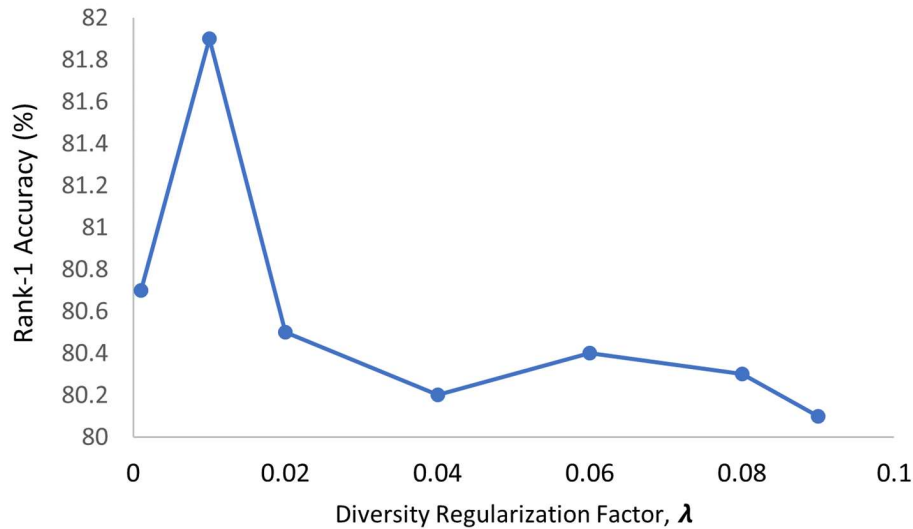

Figure 8: Deep ensemble model performance for varying hyper-parameter, $\lambda$

**Ensemble Model with Varying Backbones:** To select the baseline backbone of the independent models used in the deep ensemble model, we verify the performance of the independent models separately by varying different backbones. We train different ResNet variants for Deep Ensemble Model1 and ViT backbones for Deep Ensemble Model2. The results are presented in Fig. 9. We select ResNet-50 as the backbone of the Deep Ensemble Model1. Although ViT-H gives the best performance as shown, training time is almost twice as long for this architecture compared to ViT-L/32. Therefore, we consider ViT-L/32 as our baseline backbone for Deep Ensemble Model2 considering both time and memory efficiency.

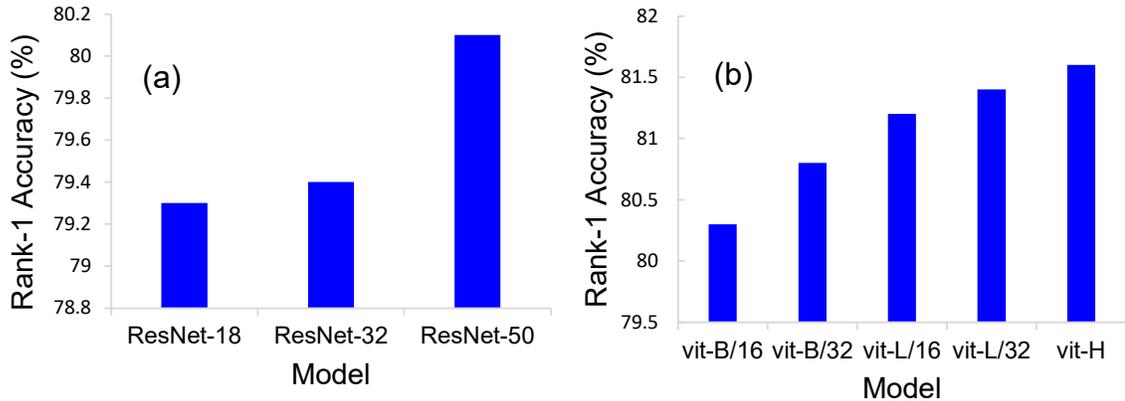

Figure 9: Performance comparison by varying the backbone network of the deep ensemble model. (a) DEM1 (b) DEM2.

**Effect of Supervision Signal:** We analyze the effect of the supervision signal and report the results in Table 5. The results show that the supervision signal generated by CNN verifier helps to increase the model performance.

Table 5: Effect of supervision signal for Occlusion Robust Token Generation.

| Module | Rank-1 | mAP |
|---|---|---|
| DEM2 w/o supervision signal | 81.3 | 82.8 |
| DEM2 w/ supervision signal | 81.4 | 83.0 |

**Influence of Human Body Mask:** In this work, we use human body part mask to learn human local body part features. Therefore, We also investigate performance of our model with varying human body mask generation strategies. Table 6 reports the performance of our model in the supervision of widely used human parsing frameworks, SCHP (48) and JPPNet (70).

Considering the re-identification accuracy, we use the masks generated by SCHP in our work.

Table 6: Performance comparison for different mask generation strategies for training DEM1.

| Module | Rank-1 | mAP |
|---|---|---|
| DEM1 JPPNet | 77.9 | 80.3 |
| DEM1 SCHP | 80.1 | 82.8 |

**Effect of Sparse Attention:** With sparsemax attention, our model improves from 79.5% to 80.1% rank-1 accuracy. Furthermore, mAP also increases by 1% as shown in Table 7.

Table 7: Performance comparison for sparsemax vs softmax attention

| Module | Rank-1 | mAP |
|---|---|---|
| DEM1 softmax | 79.5 | 81.8 |
| DEM1 sparsemax | 80.1 | 82.8 |

4.4. Discussions and Limitations

Figure 10 shows some failure cases of our model for the PRAI dataset. Based on our experimental results, we observe that rank-1 and mAP

performance for UAV-based person Re-ID remains poor. On the PRAI-1581 aerial imagery datasets, both OSNet (69) and ours can only achieve rank-1 accuracy slightly above 50%. This is within our expectation because in the less controlled situations of UAV-based imaging (i.e., high pitch and yaw angles), person Re-ID becomes more challenging due to significant loss of discriminative information, as shown in Fig. 10. From an occlusion handling perspective, we believe that the PRAI-1581 dataset can serve as a good "in the wild" test scenario - e.g., if all people wear the same hat (the occluding object), the task of identification will become impossible in this extreme situation. Simultaneous identification of multiple people from the same scene is conceptually similar to the extension of object tracking from a single to multiple people, which has not been studied in the literature. As our developed model is trained considering only single scale scenario and scale misalignment has not been investigated in this work, failure of the proposed model validates for these cases. Developing a multi-scale occlusion robust re-identification system could potentially address the limitations of the current model.

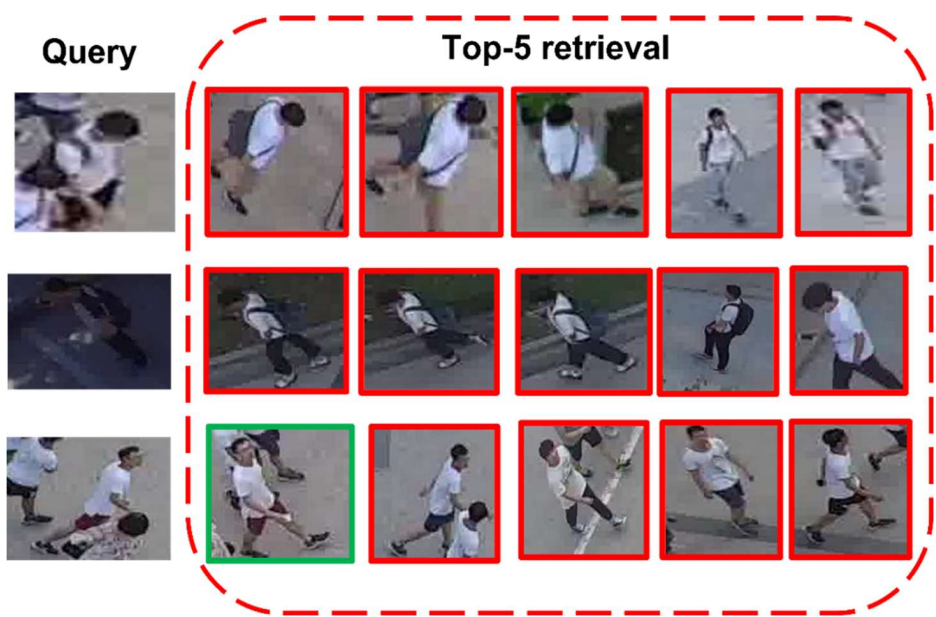

Figure 10: Failure cases of our system. Failure of image retrieval is mostly due to extreme viewing angle, low-resolution, shadow, and illumination. Red rectangular boxes denote false image retrieval, while green boxes denote the correct ones.

The other limitation of the proposed Re-ID approach lies in its problem formulation with a single query image. The video-based person Re-ID (71; 72; 73) has also received increasingly more attention in recent years. Unlike image-based approaches, video-based ReID offers new opportunities and challenges for handling occlusions because occlusion characteristics often vary in space and time. How to optimize ID performance by combining discriminative information from multiple video frames class for further extension of the orthogonal fusion strategy considered in this paper. Along this line of reasoning, we argue that a more fruitful approach to UAV-based person Re-ID is through active sensing, i.e., to actively acquire a person's biometric identity by varying camera pose (e.g., through pose-guided attention modules (74)) and distance to minimize the impact of occlusion.

## 5. CONCLUSION

This paper presents an ensemble learning-based approach to occluded person Re-ID that uses both CNN and Transformer architecture. We found that Masked AutoEncoder (MAE) reconstructed images help to boost recognition performance. Orthogonal fusion is used to generate a more discriminative feature space by combining local and global information. The proposed orthogonal fusion with occlusion handling (OFOH) method achieves a significant performance gain in comparison to state-of-the-art methods, proving the practicality of the system for an occluded person Re-ID. On both traditional and UAV-based person Re-ID datasets, our approach has demonstrated highly competent rank-1 and mAP performance compared to the current state-of-the-art approaches on five popular benchmark datasets for person Re-ID.

There are several future research directions related to occlusion handling and object Re-ID. The question of how to extend this model to aerial imagery remains an open question (75;76). How to construct a pretext task and a tracklet cluster structure for aerial imagery (cameras with high-pitch angles) deserves further study (77). Domain adaptation or cross-domain person re-ID (78) is another under-researched topic (e.g., from a fixed-position surveillance camera to a UAV-based survey). The alignment of video sequences acquired by different cameras is a long-standing open problem, in which occlusion handling is again one of the primary challenges. How to exploit the temporal redundancy in UAV-acquired video by multi-frame fusion seems a promising direction for future research. Also, the proposed method in this paper holds promise for application across diverse fields, including but not limited to microscopic image analysis (79), microbiological image analysis (80), histopathology image classification (81;82; 83), feature extraction (84; 85), cell image analysis (86; 87), video analysis (88; 89) and X-ray image analysis (90).

**CRediT authorship contribution statement**

**Syeda Nyma Ferdous:** Conceptualization, Methodology, Experimentation, Investigation, Validation, Writing – original draft.  **Xin Li:** Supervision, Writing – review & editing.